# The Statistical methods of Pixel-Based Image Fusion Techniques

Firouz Abdullah Al-Wassai[1]
Research Student, Computer Science Dept.
(SRTMU), Nanded, India
fairozwaseai@yahoo.com

N.V. Kalyankar[2]
Principal, Yeshwant Mahavidyala College
Nanded, India
drkalyankarnv@yahoo.com

Ali A. Al-Zaky[3]
Assistant Professor, Dept.of Physics, College of Science, Mustansiriyah Un.
Baghdad – Iraq.
dr.alialzuky@yahoo.com

*Abstract:* **There are many image fusion methods that can be used to produce high-resolution mutlispectral images from a high-resolution panchromatic (PAN) image and low-resolution multispectral (MS) of remote sensed images. This paper attempts to undertake the study of image fusion techniques with different Statistical techniques for image fusion as Local Mean Matching (LMM), Local Mean and Variance Matching (LMVM), Regression variable substitution (RVS), Local Correlation Modeling (LCM) and they are compared with one another so as to choose the best technique, that can be applied on multi-resolution satellite images. This paper also devotes to concentrate on the analytical techniques for evaluating the quality of image fusion (F) by using various methods including Standard Deviation (SD), Entropy(En), Correlation Coefficient (CC), Signal-to Noise Ratio (SNR), Normalization Root Mean Square Error (NRMSE) and Deviation Index (DI) to estimate the quality and degree of information improvement of a fused image quantitatively.**

*Keywords: **Data Fusion, Resolution Enhancement, Statistical fusion, Correlation Modeling, Matching, pixel based fusion.***

## I. INTRODUCTION

Satellite remote sensing offers a wide variety of image data with different characteristics in terms of temporal, spatial, radiometric and Spectral resolutions. Although the information content of these images might be partially overlapping [1], imaging systems somehow offer a tradeoff between high spatial and high spectral resolution, whereas no single system offers both. Hence, in the remote sensing community, an image with 'greater quality' often means higher spatial or higher spectral resolution, which can only be obtained by more advanced sensors [2]. However, many applications of satellite images require both spectral and spatial resolution to be high. In order to automate the processing of these satellite images new concepts for sensor fusion are needed. It is, therefore, necessary and very useful to be able to merge images with higher spectral information and higher spatial information [3]. Image fusion is a sub area of the more general topic of data fusion [4].So, Satellites remote sensing image fusion has been a hot research topic of remote sensing image processing [5]. This is obvious from the amount of conferences and workshops focusing on data fusion, as well as the special issues of scientific journals dedicated to the topic [6]. Previously, data fusion, and in particular image fusion belonged to the world of research and development. In the meantime, it has become a valuable technique for data enhancement in many applications. The term "fusion" gets several words to appear, such as merging, combination, synergy, integration … and several others that express more or less the same concept have since appeared in literature [7]. A general definition of data fusion can be adopted as fallows "Data fusion is a formal framework which expresses means and tools for the alliance of data originating from different sources. It aims at obtaining information of greater quality; the exact definition of 'greater quality' will depend upon the application" [8-10].

Many image fusion or pansharpening techniques have been developed to produce high-resolution mutlispectral images. Most of these methods seem to work well with images that were acquired at the same time by one sensor (single-sensor, single-date fusion) [11-13]. It becomes, therefore increasingly important to fuse image data from different sensors which are usually recorded at different dates. Thus, there is a need to investigate techniques that allow multi-sensor, multi-date image fusion [14]. Generally, Image fusion techniques can divided into three levels, namely: pixel level, feature level and decision level of representation [15-17]. The pixel image fusion techniques can be grouped into several techniques depending on the tools or the processing methods for image fusion procedure. This paper focuses on using statistical methods of pixel-based image fusion techniques.

This study attempts to comparing four Statistical Image fusion techniques including Local Mean Matching (LMM), Local Mean and Variance Matching (LMVM), Regression variable substitution (RVS), Local Correlation Modeling (LCM). so, This study introduces many types of metrics to examine and estimate the quality and degree of information improvement of a fused image quantitatively and the ability of this fused image to preserve the spectral integrity of the original image by fusing different sensor with different characteristics of temporal, spatial, radiometric and Spectral resolutions of TM & IRS-1C PAN images. The subsequent sections of this paper are organized as follows. Section II gives the brief overview of the related work. III covers the experimental results and analysis, and is subsequently followed by the conclusion.

## II. Statistical Methods (SM)

Different Statistical Methods have been employed for fusing MS and PAN images. They perform some type of statistical variable on the MS and PAN bands based on the local Mean Matching (LMM); on Local Mean and Variance Matching (LMVM); Regression variable substitution (RVS) and local correlation modeling (LCM) techniques applied to the multispectral images to preserve their spectral characteristics. The statistics-based fusion techniques used to solve the two major problems in image fusion – color distortion and operator (or dataset) dependency. It is different from pervious image fusion techniques in two principle ways: It utilizes the statistical variable such as the least squares; average of the local correlation or the variance with the average of the local correlation techniques to find the best fit between the grey values of the image bands being fused and to adjust the contribution of individual bands to the fusion result to reduce the color distortion.

It employs a set of statistic approaches to estimate the grey value relationship between all the input bands to eliminate the problem of dataset dependency (i.e. reduce the influence of dataset variation) and to automate the fusion process.

Some of the popular SM methods for pan sharpening are RVS, LMM, LMVM and LCM. The algorithms are described in the following sections.

To explain the algorithms through this report, Pixels should have the same spatial resolution from two different sources that are manipulated to obtain the resultant image. So, before fusing two sources at a pixel level, it is necessary to perform a geometric registration and a radiometric adjustment of the images to one another. When images are obtained from sensors of different satellites as in the case of fusion of SPOT or IRS with Landsat, the registration accuracy is very important. Therefore, resampling of MS images to the spatial resolution of PAN is an essential step in some fusion methods to bring the MS images to the same size of PAN, , thus the resampled MS images will be noted by $M_k$ that represents the set of DN of band k in the resampled MS image . Also the following notations will be used: P as DN for PAN image, $F_k$ the DN in final fusion result for band k. $\bar{M}_k$ $\bar{P}$, and $\sigma_P, \sigma_{M_k}$ Denote the local means and standard deviation calculated inside the window of size (3, 3) for $M_k$ and P respectively.

### A. The LMM and LMVM Techniques:

The general Local Mean Matching (LMM) and Local Mean Variance Matching (LMVM) algorithms to integrate two images, PAN into MS resampled to the same size as P, are given by [18,19] as follow:

**1.** The LMM algorithm:

$$F_{k(i,j)} = P_{(i,j)} \times \frac{\bar{M}_{k(i,j)(w,h)}}{\bar{P}_{(i,j)(w,h)}} \quad (1)$$

Where $F_{k(i,j)}$ is the fused image, $P_{(i,j)}$ and $M_{k(i,j)}$ are respectively the high and low spatial resolution images at pixel coordinates (i,j); $\bar{M}_{k(i,j)(w,h)}$ and $\bar{P}_{(i,j)(w,h)}$ are the local means calculated inside the window of size (w,h), which used in this study a 11*11 pixel window.

**2.** The LMVM algorithm:

$$F_{k(i,j)} = \frac{(P_{(i,j)} - \bar{P}_{(i,j)})\sigma_{M_{k(i,j)(w,h)}}}{\sigma_{P_{(i,j)(w,h)}}} + \bar{M}_{k(i,j)} \quad (2)$$

Where $\sigma$ is the local standard deviation. The amount of spectral information preserved in the fused product can be controlled by adjusting the filtering window size [18]. Small window sizes produce the least distortion. Larger filtering windows incorporate more structural information from the high resolution image, but with more distortion of the spectral values [20].

### B. Regression Variable Substitution

This technique is based on inter-band relations. Due to the multiple regressions derives a variable, as a linear function of multi-variable data that will have maximum correlation with unvaried data. In image fusion, the regression procedure is used to determine a linear combination (replacement vector) of an image channel that can be replaced by another image channel [21]. This method is called regression variable substitution (RVS) [3,11] called it a statistics based fusion, which currently implemented in the PCI& Geomatica software as special module, PANSHARP – shows significant promise as an automated technique. The fusion can be expressed by the simple regression shown in the following eq.

$$F_k = a_k + b_k \cdot P \quad (3)$$

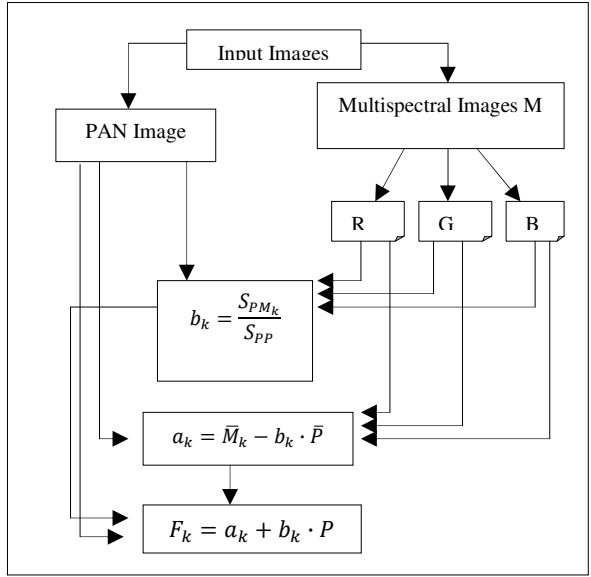

Fig. 1: Schematic of Regression Variable Substitution

The bias parameter $a_k$ and the scaling parameter $b_k$ can be calculated by a least squares approach between the resampled band MS and PAN images.

The bias parameter $a_k$ and the scaling parameter $b_k$ can be calculated by using eq. (4 & 5) between the resample bands multispectral $M_k$ and PAN band $P$ (see appendix)

$$b_k = \frac{S_{PM_k}}{S_{PP}} \quad (4)$$

Where $S_{PM_k}$ and $S_{PP}$ are the covariance between $P$ with $M_k$ of band k and the variance $P$ respectively.

$$a_k = \bar{M}_k - b_k \cdot \bar{P} \quad (5)$$

Where $\bar{M}_k$ and $\bar{P}$ are the mean of $M_k$ and $P$. Instead of computing global regression parameters $a_k$ and $b_k$ in this study, the parameter are determined in a sliding window a 5*5 pixel window was applied. the Schematic of Regression Variable Substitution is show in Fig.1

### C. Local Correlation Modeling (LCM)

The basic assumption is a local correlation, once identified between original $M^{low}$ band and downsample the PAN ($P^{low}$) should also apply to the higher resolution level. Consequently, the calculated local regression coefficients and residuals can be applied to the corresponding area of the PAN bad. The required steps to implement this technique, as given by [22 are:

1. The geometrically co-registered PAN band is blurred to match the equivalent resolution of the multispectral image.

2. The regression analysis within a small moving window is applied to determine the optimal local modeling coefficient and the residual errors for the pixel neighborhood using a single $M_k^{low}$ and the degraded panchromatic band $P^{low}$ in this study is a 11*11 pixel window.

$$M_k^{low} = a_k \times P^{low} + b_k + e_k \quad (6)$$

$$e_k = M_k^{low} - (b_k + a_k \times P^{low}) \quad (7)$$

Where $a_k$ and $b_k$ are the coefficients which can be calculated by using equations (4 & 5), $e_k$ the residuals derived from the local regression analysis of band *k*.

3. The actual resolution enhancement is then computed by using the modeling coefficients with the original PAN band *P*, where these are applied for a pixel neighborhood the dimension through the resolution difference between both images thus [22]:

$$F_k = b_k + a_k \times P + e_k \quad (8)$$

The Flowchart of Local Correlation Modeling LCM is shown in Fig. 2.

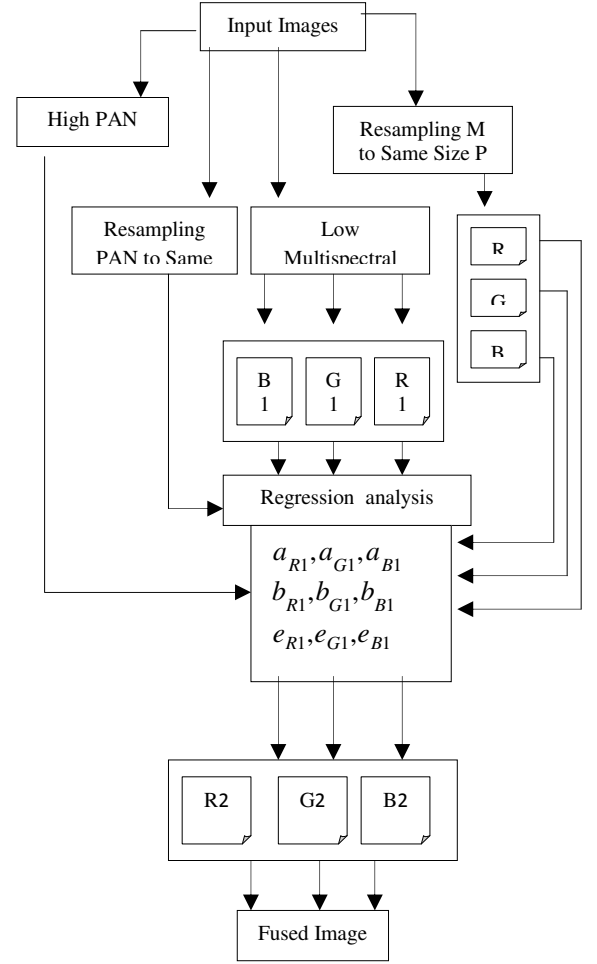

Fig. 2: Flowchart of Local Correlation Modeling

## III. Fusion image results
### i. Study Area and Datasets

In order to validate the theoretical analysis, the performance of the methods discussed above was further evaluated by experimentation. Data sets used for this study were collected by the Indian IRS-1C PAN (0.50 - 0.75 µm) of the 5.8- m resolution panchromatic band. Where the American Landsat (TM) the red (0.63 - 0.69 µm), green (0.52 - 0.60 µm) and blue (0.45 - 0.52 µm) bands of the 30 m resolution multispectral image were used in this work. Fig. 3 shows the IRS-1C PAN and multispectral TM images. The scenes covered the same area of the Mausoleums of the Chinese Tang – Dynasty in the PR China [23] was selected as test sit in this study. Since this study is involved in evaluation of the effect of the various spatial, radiometric and spectral resolution for image fusion, an area contains both manmade and natural features is essential to study these effects. Hence, this work is an attempt to study the quality of the images fused from different sensors with various characteristics. The size of the PAN is 600 * 525 pixels at 6 bits per pixel and the size of the original multispectral is 120 * 105 pixels at 8 bits per pixel, but this is upsampled to by nearest neighbor. It was used to avoid spectral contamination caused by interpolation, which does not change the data file value. The pairs of images were geometrically registered to each other.

### ii. Quality Assessment

To evaluate the ability of enhancing spatial details and preserving spectral information, some Indices including Standard Deviation (SD), Entropy(En), Correlation Coefficient (CC), Signal-to Noise Ratio (SNR), Normalization Root Mean Square Error (NRMSE) and Deviation Index (DI) of the image were used (Table 1). In the following sections, $F_k$, $M_k$ are the measurements of each the brightness values of homogenous pixels of the result image and the original multispectral image of band k, $\overline{M}_k$ and $\overline{F}_k$ are the mean brightness values of both images and are of size $n*m$. BV is the brightness value of image data $\overline{M}_k$ and $\overline{F}_k$. To simplify the comparison of the different fusion methods, the values of the En, CC, SNR, NRMSE and DI index of the fused images are provided as chart in Fig. 4.

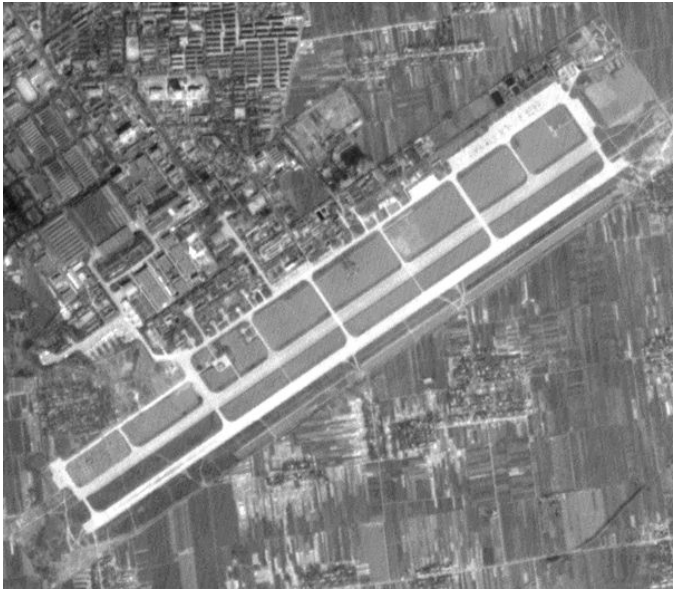

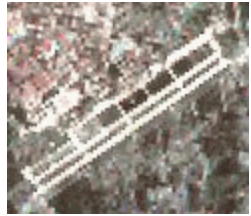

Fig.3: The Representation of Original Panchromatic and Multispectral Images

| Equation |
|---|
| $\sigma = \sqrt{\dfrac{\sum_{i=1}^{m}\sum_{j=1}^{n}(BV(n,m)-\mu)^2}{m \times n}}$ |
| $CC = \dfrac{\sum_i^n \sum_j^m (F_k(i,j) - \overline{F}_k)(M_k(i,j) - \overline{M}_k)}{\sqrt{\sum_i^n \sum_j^m (F_k(i,j) - \overline{F}_k)^2}\sqrt{\sum_i^n \sum_j^m (M_k(i,j) - \overline{M}_k)^2}}$ |
| $En = -\sum_{0}^{I-1} P(i) \log_2 P(i)$ |
| $DI_k = \dfrac{1}{nm}\sum_i^n \sum_j^m \dfrac{|F_k(i,j) - M_k(i,j)|}{M_k(i,j)}$ |
| $SNR_k = \sqrt{\dfrac{\sum_i^n \sum_j^m (F_k(i,j))^2}{\sum_i^n \sum_j^m (F_k(i,j) - M_k(i,j))^2}}$ |

$$NRMSE_k = \sqrt{\frac{1}{nm*255^2}\sum_i^n\sum_j^m(F_k(i,j)-M_k(i,j))^2}$$

## IV. Results And Discussion

From table2 and Fig. 4 shows those parameters for the fused images using various methods. It can be seen that from Fig. 4a and table2 the SD results of the fused images remains constant for RVS. According to the computation results En in table2, the increased En indicates the change in quantity of information content for radiometric resolution through the merging. From table2 and Fig.4b, it is obvious that En of the fused images have been changed when compared to the original multispectral. In Fig.4c and table2 the maximum correlation values were for RVS and LCM also, the maximum results of SNR were for RVS and LCM. The results of $SNR$, NRMSE and DI appear changing significantly. It can be observed, from table2 with the diagram of Fig. 4d & Fig. 4e, that the results of SNR, NRMSE & DI of the fused image, show that the RVS method gives the best results with respect to the other methods indicating that this method maintains most of information spectral content of the original multispectral data set which gets the same values presented the lowest value of the NRMSE and DI as well as the higher of the CC and SNR. Hence, the spectral quality of fused image RVS technique is much better than the others. In contrast, it can also be noted that the LMM and LMVM images produce highly NRMSE & DI values indicating that these methods deteriorate spectral information content for the reference image. By comparing the visual inspection results, it can be seen that the experimental results overall method During this work, it was found that the RVS in Fig.5c has a higher resolution compared to the other results. RVS method gives the best results with respect to the other methods. Fig.3. shows the original images and Fig.5 the fused image results.

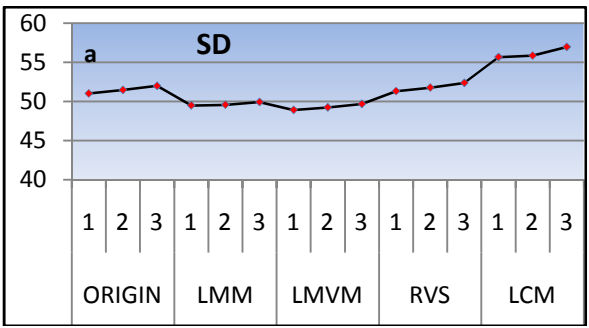

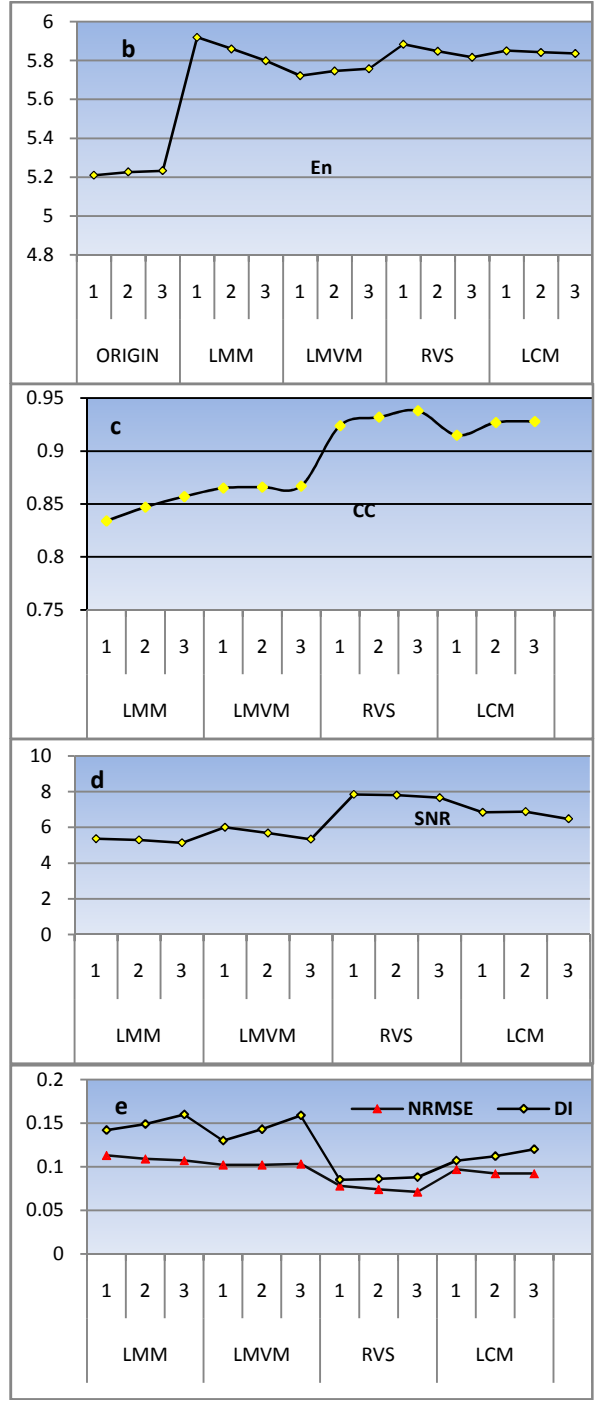

**Fig. 4:** Chart Representation of SD, En, CC, SNR, NRMSE & DI of Fused Images

Table 2: Quantitative Analysis of Original MS and Fused Image Results
Through the Different Methods

| Method | Band | SD | En | SNR | NRMSE | DI | CC |
|---|---|---|---|---|---|---|---|
| ORIGIN | 1 | 51.018 | 5.2093 | | | | |
| | 2 | 51.477 | 5.2263 | | | | |
| | 3 | 51.983 | 5.2326 | | | | |
| LMM | 1 | 49.5 | 5.9194 | 5.375 | 0.113 | 0.142 | 0.834 |
| | 2 | 49.582 | 5.8599 | 5.305 | 0.109 | 0.149 | 0.847 |
| | 3 | 49.928 | 5.7984 | 5.146 | 0.107 | 0.16 | 0.857 |
| LMVM | 1 | 48.919 | 5.7219 | 6.013 | 0.102 | 0.13 | 0.865 |
| | 2 | 49.242 | 5.746 | 5.69 | 0.102 | 0.143 | 0.866 |
| | 3 | 49.69 | 5.7578 | 5.349 | 0.103 | 0.159 | 0.867 |
| RVS | 1 | 51.323 | 5.8841 | 7.855 | 0.078 | 0.085 | 0.924 |
| | 2 | 51.769 | 5.8475 | 7.813 | 0.074 | 0.086 | 0.932 |
| | 3 | 52.374 | 5.8166 | 7.669 | 0.071 | 0.088 | 0.938 |
| LCM | 1 | 55.67 | 5.85 | 6.854 | 0.097 | 0.107 | 0.915 |
| | 2 | 55.844 | 5.842 | 6.891 | 0.092 | 0.112 | 0.927 |
| | 3 | 56.95 | 5.8364 | 6.485 | 0.092 | 0.12 | 0.928 |

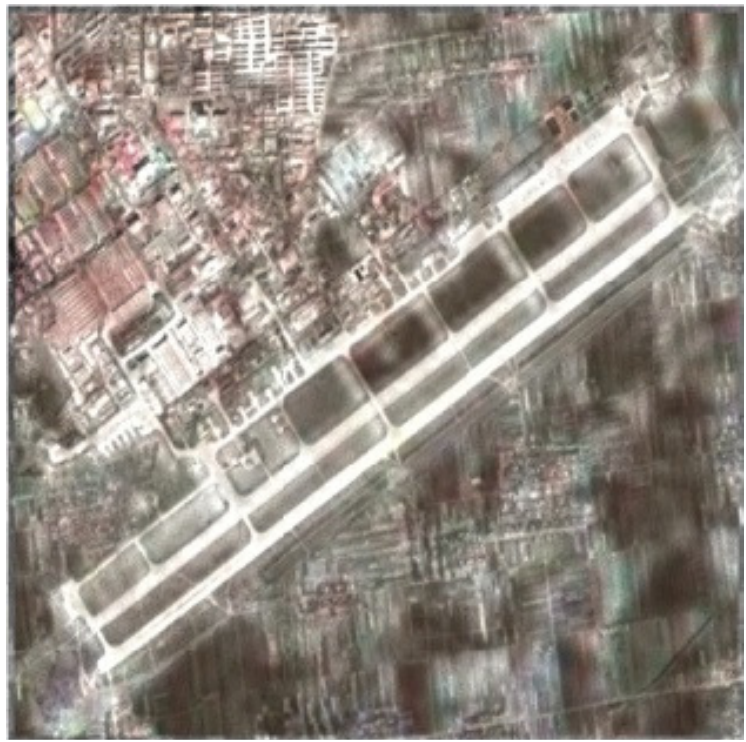

Fig.5a: The Representation of Fused Images (LMM)

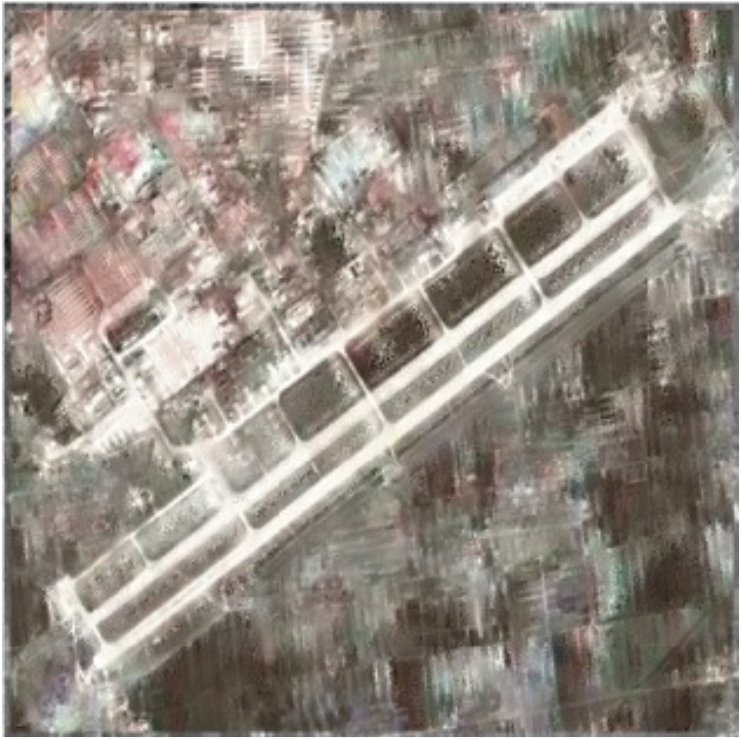

Fig.5b: The Representation of Fused Images (LMVM)

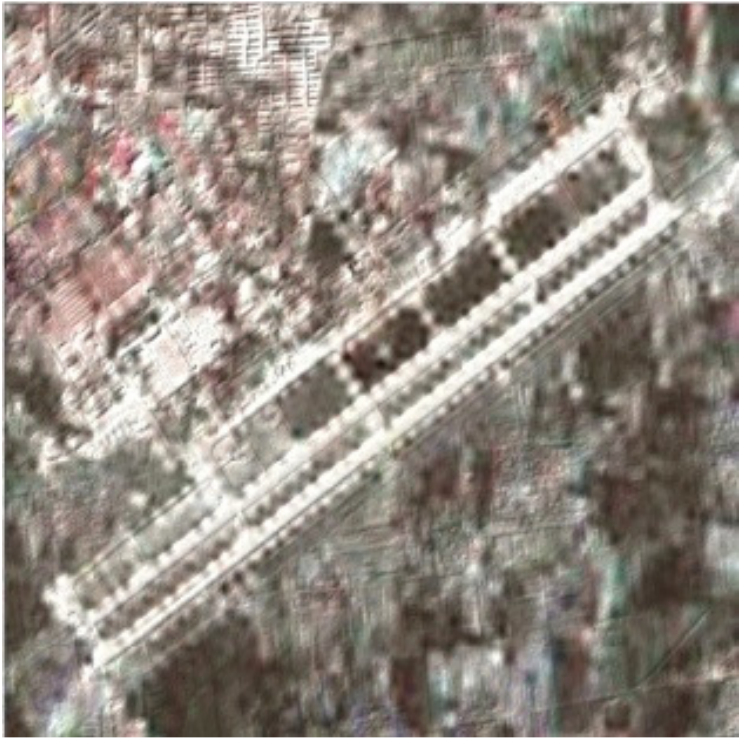

Fig.5c: The Representation of Fused Images(RVS)

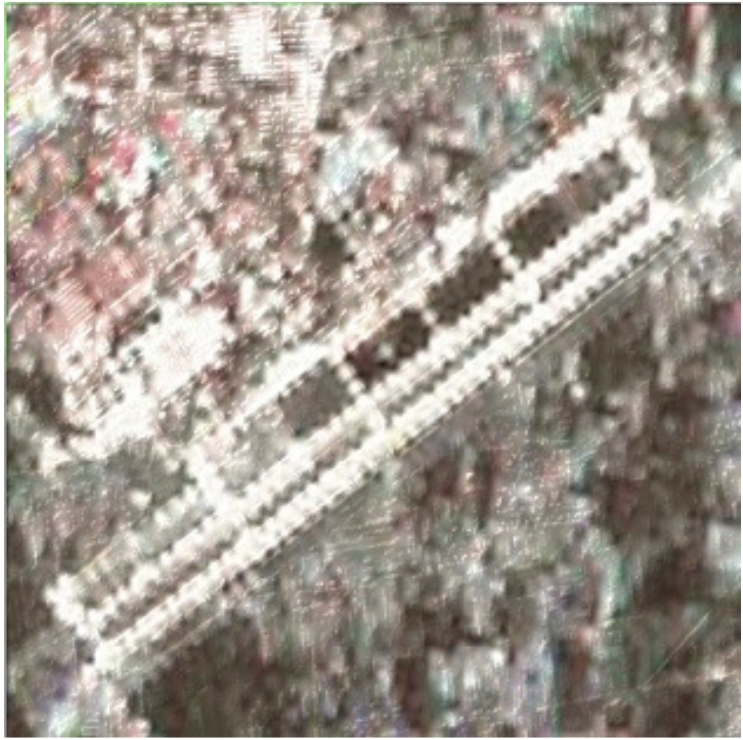

Fig.5d: The Representation of Fused Images(LCM)

Fig.5: The Representation of Fused Images

## V. Conclusion

In this paper, the comparative studies undertaken by statistical methods based pixel image fusion techniques as well as effectiveness based image fusion and the performance of these methods have been studied. The preceding analysis shows that the RVS technique maintains the spectral integrity and enhances the spatial quality of the imagery.

The use of the RVS based fusion technique could, therefore, be strongly recommended if the goal of the merging is to achieve the best representation of the spectral information of multispectral image and the spatial details of a high-resolution panchromatic image because it utilizes the statistical variable of the least square to find the best fit between the grey values of the image bands being fused and to adjust the contribution of individual bands to the fusion result to reduce the color distortion as well as employs a set of statistic approaches to estimate the grey value relationship between all the input bands to eliminate the problem of dataset dependency. Also, the analytical technique of DI is much more useful for measuring the spectral distortion than NRMSE since the NRMSE gave the same results for some methods; but the DI gave the smallest different ratio between those methods, therefore , it is strongly recommended to use the DI because of its mathematical more precision as quality indicator.

## VI. AKNOWLEDGEMENTS

The Authors wish to thank our friend Fatema Al-Kamissi at University of Ammran( Yemen) for her suggestion and comments.

**AUTHORS**

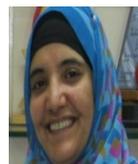

*Firouz Abdullah Al-Wassai*. Received the B.Sc. degree in, Physics from University of Sana'a, Yemen, Sana'a, in 1993. The M.Sc.degree in, Physics from Bagdad University , Iraq, in 2003, Research student.Ph.D in the department of computer science (S.R.T.M.U), India, Nanded.

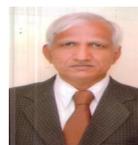

*Dr. N.V. Kalyankar*, Principal,Yeshwant Mahvidyalaya, Nanded(India) completed M.Sc.(Physics) from Dr. B.A.M.U, Aurangabad. In 1980 he joined as a leturer in department of physics at Yeshwant Mahavidyalaya, Nanded. In 1984 he completed his DHE. He completed his Ph.D. from Dr.B.A.M.U. Aurangabad in 1995. From 2003 he is working as a Principal to till date in Yeshwant Mahavidyalaya, Nanded. He is also research guide for Physics and Computer Science in S.R.T.M.U, Nanded. 03 research students are successfully awarded Ph.D in Computer Science under his guidance. 12 research students are successfully awarded M.Phil in Computer Science under his guidance He is also worked on various boides in S.R.T.M.U, Nanded. He is also worked on various bodies is S.R.T.M.U, Nanded. He also published 30 research papers in various international/national journals. He is peer team member of NAAC (National Assessment and Accreditation Council, India ). He published a book entilteld "DBMS concepts and programming in Foxpro". He also get various educational wards in which "Best Principal" award from S.R.T.M.U, Nanded in 2009 and "Best Teacher" award from Govt. of Maharashtra, India in 2010. He is life member of Indian "Fellowship of Linnean Society of London(F.L.S.)" on 11 National Congress, Kolkata (India). He is also honored with November 2009.

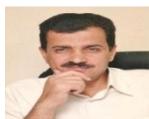

*Dr. Ali A. Al-Zuky*. B.Sc Physics Mustansiriyah University, Baghdad , Iraq, 1990. M Sc. In1993 and Ph. D. in1998 from University of Baghdad, Iraq. He was supervision for 40 postgraduate students (MSc. & Ph.D.) in different fields (physics, computers and Computer Engineering and Medical Physics). He has More than 60 scientific papers published in scientific journals in several scientific conferences.